\documentclass[runningheads]{llncs}

\usepackage[T1]{fontenc}
\usepackage{graphicx}
\usepackage{listings}
\usepackage{hyperref}

\usepackage{color}

\begin{document}
\title{
    An Architecture for Unattended Containerized (Deep) Reinforcement Learning with Webots
}
\titlerunning{Unattended Containerized (Deep) Reinforcement Learning}

\author{
    Tobias Haubold \and Petra Linke
}
\authorrunning{T. Haubold, P. Linke}

\institute{University of Applied Sciences Zwickau, Germany \\
    \email{toh@fh-zwickau.de},
    \email{petra.linke@fh-zwickau.de}
\\ 
\vspace{3mm}
    November 2, 2023
}

\maketitle
\begin{abstract}
As data science applications gain adoption across industries, the tooling landscape matures to facilitate the life cycle 
of such applications and provide solutions to the challenges involved to boost the productivity of the people involved. 
Reinforcement learning with agents in a 3D world could still face challenges: the knowledge required to use a 
simulation software as well as the utilization of a standalone simulation software 
in unattended training pipelines.
In this paper we review tools and approaches to train reinforcement learning agents for robots in 3D worlds with respect 
to the robot Robotino and argue 
that the separation of the simulation environment for creators of virtual worlds and the model development environment for data 
scientists is not a well covered topic. Often both are the same and data scientists require knowledge of the simulation software 
to work directly with their APIs. Moreover, sometimes creators of virtual worlds and data scientists even work on the same files. 
We want to contribute to that topic by describing an approach where data scientists don't require knowledge about the simulation 
software. Our approach uses the standalone simulation software Webots, the Robot Operating System 
to communicate with simulated robots as well as the simulation software itself and 
container technology to separate the simulation from the model development environment. 
We put emphasize on the APIs the data scientists work with 
and the use of a standalone simulation software in unattended training pipelines. 
We show the parts that are specific to the Robotino and the robot task to learn.

\end{abstract}

\counterwithin{lstlisting}{section}

\section{Motivation}

Over the last years there were considerable progress in the field of reinforcement learning. 
Several new algorithms including the dqn agent\cite{dqn} were published showing remarkable results for problem solving. 
The gym library\cite{gym} was released and established a programming interface together with the 
agent-environment-loop\cite{rl-book} showing how to interface single agent reinforcement learning environments in 
an agent agnostic way. 
With MuJoCo\cite{mujoco} an advanced physics simulator was made available as open source to facilitate research and 
development. 

In the same time containerized approaches matured in the field of infrastructure. Containers are a standard unit of 
software that packages software code and its dependencies together with system libraries and tools as well as settings. 
So applications can be setup quickly in different computing environments and run quickly and reliably. 
All container run isolated from each other by a container runtime and share the host operating system. 

The development, deployment and life cycle of data science applications in the industry matured over the last years 
and shaped the tooling landscape. 
An important part is the definition of a pipeline, usually in the form of a directed acyclic graph, and the use of 
job processing systems to train models in an unattended, reproducible and scalable manner. 
In that regard reinforcement learning setups still face the challenge that data scientists need knowledge 
about the simulation software. If they strive to use a standalone simulation software 
for agents in a 3D world, running training sessions unattended might be another challenge. 
Although that is straight forward with libraries like \lstinline{MuJoCo}\cite{mujoco}, other approaches require 
to define the learning environment in the simulation software like \lstinline{Unity ml-agents}\cite{unity.ml-agents} or 
move the training of agents into the simulation software like \lstinline{Deepbots}\cite{deepbots}. 
Both approaches require that data scientists have some familiarity with the simulation software and their APIs. 
We like to contribute to that topic with an approach where data scientists don't require knowledge about the simulation software. 
Instead they use Python based APIs 
according the Facade pattern\cite[pp. 185ff]{design-pattern-book} 
and the simulation software is started on demand. 
Under the hood an established communication mechanism is used to interact with the simulation.

Our research task at hand involves the robot Robotino in logistical settings. 
We strive for an approach that is applicable to other robots as well, 
enables expansion of simple logistical settings into more complex sceneries and 
facilitates a structured teamwork with clear responsibilities between the typical roles in data science applications 
beside the subject matter experts: 
data scientists including method developers, creators of virtual worlds and the infrastructure team. 

We first review related work in section \ref{sec:related}, discuss the architecture and implementation details in 
section \ref{sec:approach}, outline its application on a sample task involving the Robotino in section \ref{sec:example}, 
discuss current limitations in section \ref{sec:limitations} and summarize our experience in section \ref{sec:summary}.

\section{Related Work}
\label{sec:related}

\subsection{Robotino Sim Pro}
Festo, the company behind the robot Robotino\footnote{
    Robotino 4 product website: \url{https://ip.festo-didactic.com/InfoPortal/Robotino/Overview/EN/index.html}
} provides Robotino Sim Pro\footnote{
    Robotino Sim Pro product website: \url{https://www.festo.com/at/en/p/robotino-sim-professional-id_PROD_DID_567230/?page=0}
}, a simulation environment tailored to the Robotino, on a commercial basis. There is a demo version\footnote{
    Robotino Sim Demo website: \url{https://ip.festo-didactic.com/InfoPortal/Robotino/Software/Simulation/EN/index.html}
} available for evaluation purpose.
The most notable shortcoming is that it only runs on Microsoft Windows and supports the versions 2000, XP, Vista and 7.
Since Microsoft dropped support of Windows 7 in January 2020 and for the other versions even earlier it is no longer safe 
to operate it.

With regard to the simulated hardware the support of the additional modules\footnote{
    Robotino Modules product website: \url{https://ip.festo-didactic.com/InfoPortal/Robotino/Hardware/Modules/EN/index.html}
} Festo provides for the Robotino is limited. 
E. g. the laser range finder is supported but the forklift and the electric gripper is not.

\subsection{Robot Operating System (ROS)}
The field of robot control is dominated by \lstinline{ROS}\cite{ROS} for many years. 
It is an open-source robotics middleware rather than an operating system and provides a set of software frameworks and 
libraries for robot software development. It is well-established in the robotics industry and in research. 

With regard to our task it provides a heterogenous network that consists of a number of nodes, possibly located on 
different hosts. In terms of communication mechanisms it provides synchronous communication with services and 
asynchronous communication with a publisher-subscriber model. For data transfer it provides 
standardized data structures, e. g. for images, as well as the possibility to define custom data structures.

We are going to use ROS to communicate with the real Robotino as well as the virtual Robotino, see \ref{sec:robotino}.

\subsection{gymnasium}
In the area of reinforcement learning an important topic is to decouple the learning algorithms from the environments. 
The company \emph{Open AI} open sourced the library \lstinline{gym}\cite{gym} providing the 
programming interface \lstinline{Env} as an 
agent agnostic interface for single agent reinforcement learning environments. That interface is a well-established approach to 
separate those concerns. 
The library \lstinline{gym} is no longer maintained. All future development takes place in the 
library \lstinline{gymnasium}\cite{gymnasium} which serves as a drop-in replacement for \lstinline{gym}.

The basic concept behind the \lstinline{Env} interface is the \emph{agent-environment loop}\cite[p. 48]{rl-book}. 
An agent (or policy) receives an initial observation (or state) from an environment. Based on the observation the agent 
choose to perform an action in the environment. The agent receives the new observation and a reward from the environment. 

The \lstinline{Env} interface is a well-established interface for reinforcement learning environments. We are going to use 
that interface to describe our sample environment in section \ref{sec:env}.

\subsection{MuJoCo}
Reinforcement learning environments involving robots usually leverage a physics engine to model behavior. 
\lstinline{MuJoCo}\cite{mujoco} stands for \emph{Multi-Joint dynamics with Contact} and is a physics engine that aims to 
facilitate research and development in robotics and other areas. It was a commercial physics engine until Google DeepMind 
acquired it in October 2021 and open-sourced it in 2022.

There are a couple of sample environments in \lstinline{gymnasium} that use \lstinline{MuJoCo}. They all provide simple 3D 
visualizations based on OpenGL. With its Python API it can be easily used as a library from Python code and a plugin enables 
the game engine \lstinline{Unity} to use \lstinline{MuJoCo} as physics engine. 

With regard to our task there is no work available involving the robot Robotino. 
The provided Python API is an easy way to interact with the simulator and the robot but 
remains specific to the library. 
With regard to industrial applications, the intended use appears to be as physics plugin of the game engine \lstinline{Unity}. 
The decision about the robot simulation software was a tough one but we decided to continue our work with \lstinline{Webots} 
as outlined in \ref{sec:webots} and \ref{sec:unity-ml-agents}. 
Note that the presented approach is not required if \lstinline{MuJoCo} is used solely or if \lstinline{Unity} together with 
\lstinline{ml-agents} (see section \ref{sec:unity-ml-agents}) is used as simulation software 
with \lstinline{MuJoCo} as physics engine.

\subsection{Webots}
\label{sec:webots}
\lstinline{Webots}\cite{webots} is an open source mobile robot simulation software maintained by Cyberbotics Ltd. 
It provides a couple of robots\footnote{
    Overview of Robots in Webots: \url{https://webots.cloud/proto?keyword=robot}
} and sample worlds together with sample controllers that illustrate how to control the robots. 
Webots supports a so called \emph{headless} mode to run without graphical user interface that is suited for batch processing on 
a server. It supports different programming languages including Python and integrates with ROS.

With regard to our task it provides a model for the robot Robotino, a controller and a sample world with it. 
The integration of ROS offers the possibility to control the simulation itself. The default ROS controller for robots provides 
the robots with their properties and abilities on the ROS network as nodes. Thus to interface the simulation and robots 
only the client code needs to be developed. 
Since there is a ROS API for the real Robotino, both the real and the virtual Robotino could be controlled with the same 
communication mechanism. 

Based on the standardized communication mechanism with ROS, the 3D world simulator and the perspective to leverage all 
available robots in their sample worlds for reinforcement learning
we decided to use \lstinline{Webots}.

\subsection{Deepbots}
The deep reinforcement learning framework Deepbots\cite{deepbots} combines the simulation environment Webots with Open AI gym,
the standard API for reinforcement learning environments. It provides a class structure that enables the communication as well as
the data transport between the Webots supervisor controller (controls the simulation), the robot controller and an 
Open AI gym environment.
It supports the data formats CSV and JSON for data transport between the robot controllers and the gym environment.

Deepbots makes use of the programming language support of Webots to write the supervisor and robot controllers in Python 
source code. It uses Webots as its main user interface for reinforcement learning: the user starts Webots, loads a world and 
starts the simulation to train reinforcement learning agents. The robot controllers are started as Python subprocesses by Webots. 
That way Webots is used as training environment and training sessions run entirely inside Webots, which requires some familiarity 
of data scientists with Webots and its APIs.

The work presented in this paper relocates the \emph{training} outside of Webots and separates the training from the simulation 
environment. By doing so the training environment becomes the \emph{active} part and automatically starts Webots if needed 
and loads the required world too, without user interaction. 
The trigger to train reinforcement learning agents could be a user or any software system that 
starts a python script or executes a jupyter notebook. 

Compared with Deepbots the proposed approach comes with more complexity and error-proneness. But allows on the other hand for
higher automation and scalability. Training sessions could run unattended in a pipeline as a job, e. g. within a continuous
integration or job processing system. Data scientists don't require any knowledge about Webots.

\subsection{Unity ML-Agents Toolkit}
\label{sec:unity-ml-agents}
Unity is one of the most popular game engines. Unity Technologies, the company behind the game engine, started the 
open source project Unity Machine Learning Agents Toolkit (ml-agents)\cite{unity.ml-agents} that 
enables the game engine to be used to train reinforcement learning agents.
It provides example environments\footnote{
    Unity ml-agents example learning environments: 
    \url{https://github.com/Unity-Technologies/ml-agents/blob/develop/docs/Learning-Environment-Examples.md}
} together with implementations of algorithms based on PyTorch\footnote{
    A machine learning framework for Python like tensorflow: \url{https://pytorch.org/}.
} and integrations for gym and TensorBoard\footnote{
    A visualization toolkit for machine learning experiments, e. g. metrics, histograms, etc. 
    Website: \url{https://www.tensorflow.org/tensorboard}.
}.

The ml-agents toolkit solves many challenges the presented approach has faced including the current implementation limitation 
to use multiple simulation instances (see section \ref{sec:limitations}), and makes the Unity engine 
accessible for reinforcement learning. 
However, there are fundamental differences between the approaches. Most notable ml-agents extends the Unity SDK with 
reinforcement learning specific concepts like an \lstinline{Agent}. The agent is defined in the virtual world together 
with a method that returns the agents observation, an event handlers that define its behavior when an episode starts, 
what to do when an action is received, the reward that the agent receives for the action and the conditions under which the 
episode ends. That means virtual worlds must be specifically created to be used for reinforcement learning. 

The python package with the gym integration provides the learning environments defined in Unity as gym environments to be used 
with other or custom reinforcement learning algorithms. 

The approach presented in this paper does not extend Webots and does not define reinforcement learning specific concepts or 
settings in the virtual world (see section \ref{sec:world}). Instead these definitions are done in Python source code with 
the definition of the gymnasium environment, see section \ref{sec:env}.

Compared with ml-agents, ml-agents requires that data scientists are familiar with Unity, ml-agents and the 
programming language C\# to define the environment together with the agent and the reward function in the Unity project. 
With the proposed approach data scientists don't require any knowledge about Webots but use Python APIs 
(see section \ref{sec:webots}, \ref{sec:control} and \ref{sec:robotino}) to implement 
gymnasium environments directly. Having the gymnasium environment decoupled from the simulation environment allows to use the 
virtual world for several gymnasium environments, e. g. for different scale of difficulty.

With regard to our task there is no work available involving the robot Robotino. Without a doubt Unity would have visually 
more appealing simulations.

\subsection{Nvidia Omniverse and Isaak SDK}
Nvidia provides a physics engine and robot simulator with \lstinline{Nvidia Omniverse} and \lstinline{Isaak SDK}. 
By the time of evaluation both were in an early development phase and undergoing a lot of changes. There is a list of 
available robots\footnote{
    Environments and robots included in Isaac Sim: \url{https://docs.omniverse.nvidia.com/isaacsim/latest/reference_assets.html}
} but the Robotino is not included. 
In addition the future software license regarding the terms and conditions for use were unclear and thus we decided to take 
the Nvidia tools out of consideration.

\subsection{Docker and Container}
Container technology in general and Docker\cite{docker} in particular enable to package applications together with their 
dependencies into portable images from which container could be started that run isolated from each other. 
Virtual networks enable isolated communication between containers.

Docker is a popular and established approach that provides dockerfiles to automate the build of container images, an image 
registry to distribute images and the compose file format to define and run multi-container applications. 
It has support to run container on multiple hosts but its container orchestration capabilities are not comparable with kubernetes.

We use docker to bundle the necessary software into images and use docker compose files to make the setup of 
execution environments with multiple container services (illustrated in figure \ref{fig:setup}) easy and fast. 
With virtual networks the setups to train reinforcement learning agents are replicable and scalable, even on a single machine.

\section{Proposed Approach}
\label{sec:approach}
\begin{figure}[htb]
\includegraphics[width=\textwidth]{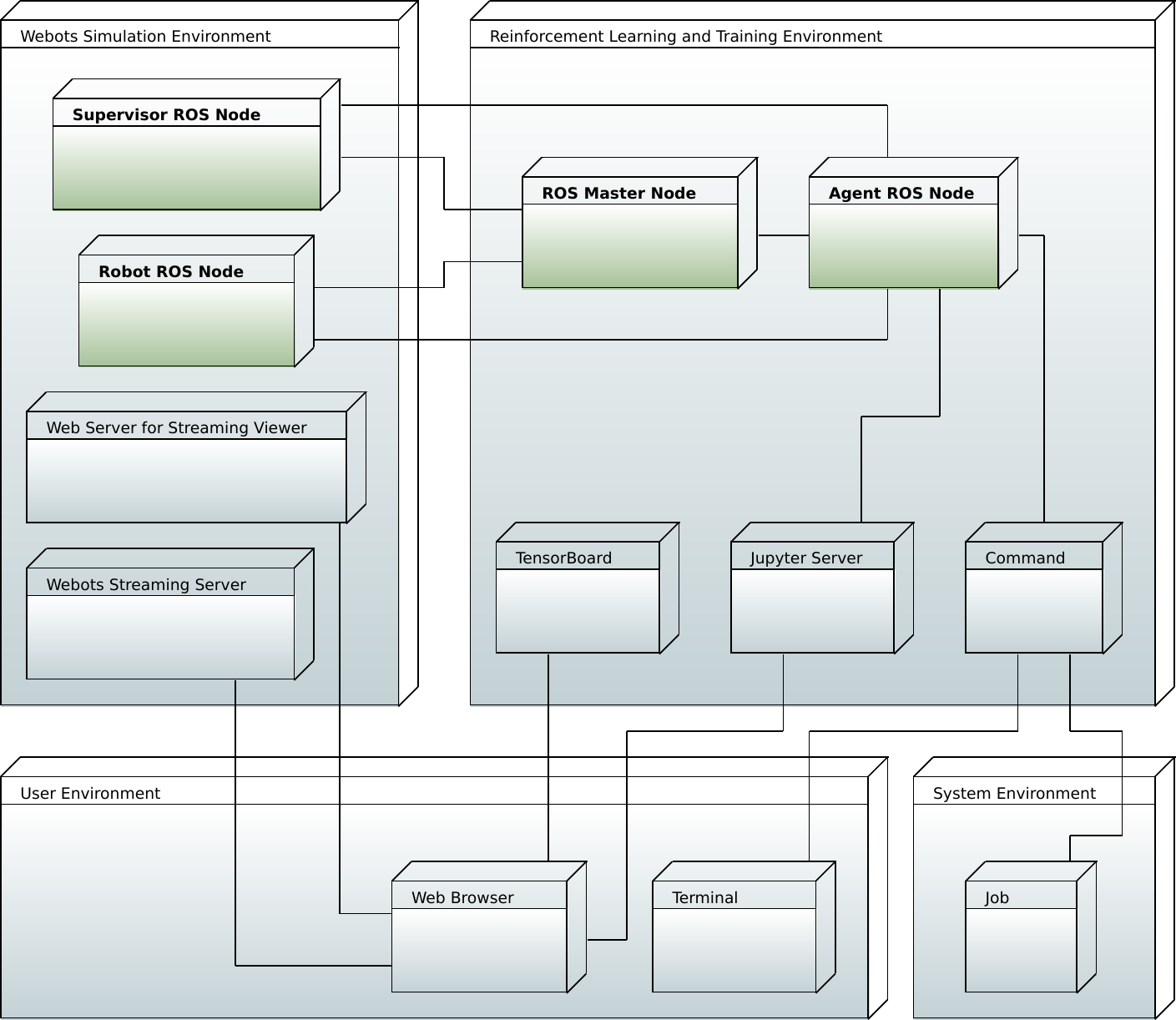}
\caption{Top-down perspective on the setup} \label{fig:setup}
\end{figure}

From a domain perspective our goal is to use reinforcement learning to train agents for the robot Robotino in the 
Webots simulation environment without the need for human interaction or even initiation. 
Figure \ref{fig:setup} illustrates the setup from a top-down perspective.
Essentially there are two containers: one with the Webots simulation environment and the other with the training environment. 
The nodes with bold names represent the ROS network. 
Both the user and system environments illustrate possible use cases of the container setup. 
\begin{description}
    \item[Webots Simulation Environment] is a container service running the Webots simulator. 
    It is started by the reinforcement learning environment when needed. When a world is loaded Webots starts the 
    default ROS controller for each robot in the world and the respective nodes become available on the ROS network. 
    An optional web server provides images and videos captured from the simulation as well as the Webots streaming viewer. 
    The latter could be used to connect to the Webots Streaming Server to observe the simulation.
    \item[Reinforcement Learning Environment] is a container service that runs the ROS master as well as the agent ROS node and
    provides all machine learning frameworks and libraries. The agent ROS node acts as a client and controls the simulation 
    using the supervisor and the robot ROS nodes of the Webots simulation environment. A running TensorBoard could be used to 
    monitor, view and evaluate training sessions. The container is started by running the jupyter sever or any other command.
    \item[User Environment] is any device with at least a web browser to provide access to a jupyter lab or notebook 
    environment provided by the jupyter server. It allows data scientists to work interactively with jupyter notebooks and the 
    simulation, mainly to experiment and to setup training sessions. With access to a container manager a terminal could be used 
    to start a training session manually. 
    \item[System Environment] is any environment that runs a continuous integration system, job scheduler or processing system 
    or workflow manager with access to a container manager where any job or workflow step could start a training session, 
    e. g. on changes in a code repository.
\end{description}
A command starts a reinforcement learning environment container and runs a training session by executing a 
Python script, jupyter notebook or alike. 

The containerized approach enables but doesn't mandate a distributed architecture. The Webots container should run on a system
with a graphics card and a graphical user interface to use hardware acceleration for simulation. On systems without OpenGL
the simulation uses software rendering and runs much slower. The reinforcement learning environment may run on a system with
an accelerator like a graphics or tensor processing unit.

All environments could be on the same device. 
If the user uses the device for other tasks as well Webots provides the command line argument \lstinline{--minimize} 
to reduce disruptions of the user workflow by minimizing the graphical user interface on startup.

Note that users who follow the one process per container paradigm more rigidly could use more containers for the services.

Next, we have a closer look at the building blocks to run that setup.

\subsection{Webots Facade}
\label{sec:webots.facade}
The reinforcement learning environment starts the Webots simulator on demand.
That task requires access to the container manager or orchestration service. There are python libraries for both docker and 
kubernetes offering an API for container services.

In order to hide the complexities of container management and their APIs from users, we provide a class according the 
Facade pattern that wraps container APIs and encapsulates the required functionality.
The API is shown in listing \ref{lst:webots}.
\begin{lstlisting}[caption={Facade API to Start and Stop Webots}, label={lst:webots}, captionpos=b, language=Python]
class Webots:
    def __init__(self, supervisor_node_name, world_path):
        pass
    def start(self): pass
    def stop(self): pass
    def __call__(self): pass
    def available(self): pass
    def run(self, callback, restart=False): pass
    def __enter__(self): pass
    def __exit__(self, exc_t, exc_o, traceback): pass
\end{lstlisting}
The initializer takes two arguments: the ROS node name of the supervisor and the file path of the world to load.
Optionally Webots specific command line options could be exposed as initializer arguments as well.

Besides \lstinline{start()} and \lstinline{stop()} to start and stop Webots we provide a restart functionality with the
call operator \lstinline{()}. The method \lstinline{available()} checks if Webots is currently running and 
returns \lstinline{True} or \lstinline{False} respectively.

One important design decision is how to stop the Webots simulator. Basically we could just kill the process or stop the whole
container. Instead we opted for the possibility to stop Webots in a controlled way using its supervisor API. This is one reason 
for the supervisor ROS node name as initializer argument.

Another important design decision is how to implement the \lstinline{available()} method. Basically we could track the 
\lstinline{start()} and \lstinline{stop()} calls or rely on the running process with some additional time given for 
startup and initialization. However, we implement that check based on the availability of the supervisor node on the ROS network. 
With regard to robustness of network communications that kind of implementation is more suited. This is another reason 
for the ROS node name of the supervisor as initializer argument.

During our work we experienced an odd behavior of Webots, possibly related the version we used. After starting a simulation 
Webots uses the cpu cores heavily, not just when performing simulation steps but also in between them when there is nothing 
to simulate. To prevent the waste of resources we make use of the simulation modes Webots offers: we switch to the simulation 
mode \emph{as fast as possible} when Webots is required to run the simulation and then we switch back to the simulation mode 
\emph{pause}.
The facade provides two possible ways to switch the simulation mode: the function \lstinline{run()} 
and the context manager provided by the special functions \lstinline{enter()} and \lstinline{exit()} that enable the use of the 
python \lstinline{with} statement. 
Both approaches follow the decorator pattern\cite[pp. 175ff]{design-pattern-book}. 
The function \lstinline{run()} additionally offers the possibility to restart Webots before the 
callback function is invoked.

\subsection{Simulation and Robot Control as well as Data Transfer}
\label{sec:control}
Having multiple processes that need to communicate requires an inter-process communication mechanism, specifically a protocol and
a data format for data exchange. ROS provides a solid base: naming and location services for nodes, synchronous communication with
services, asynchronous communication with publishers and subscribers based on topics, libraries with standardized data structures
for common use cases and mechanisms to define custom data structures. Webots with its first class support of ROS provides the whole
simulation side of the communication.

\begin{figure}[htb]
    \includegraphics[width=\textwidth]{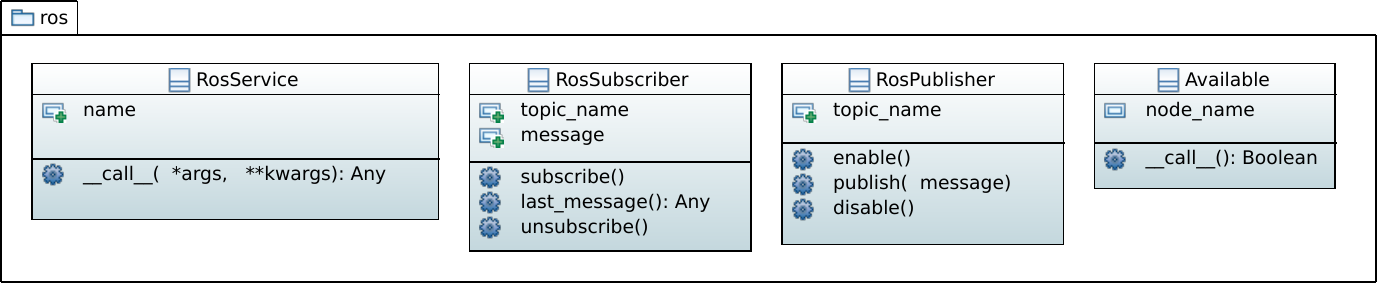}
    \caption{Facade Classes for Communication with ROS} \label{fig:ros}
\end{figure}

For the reinforcement learning side we essentially need two things: a counterpart for the supervisor ros node to control the 
simulation and a counterpart for the robot ros node to obtain sensor values and control its actuators. 
From a software design perspective it made sense to wrap ROS communication concepts and required functionality into classes 
according the Facade pattern with a straight forward API shown in figure \ref{fig:ros}.

\begin{figure}[htb]
    \includegraphics[width=\textwidth]{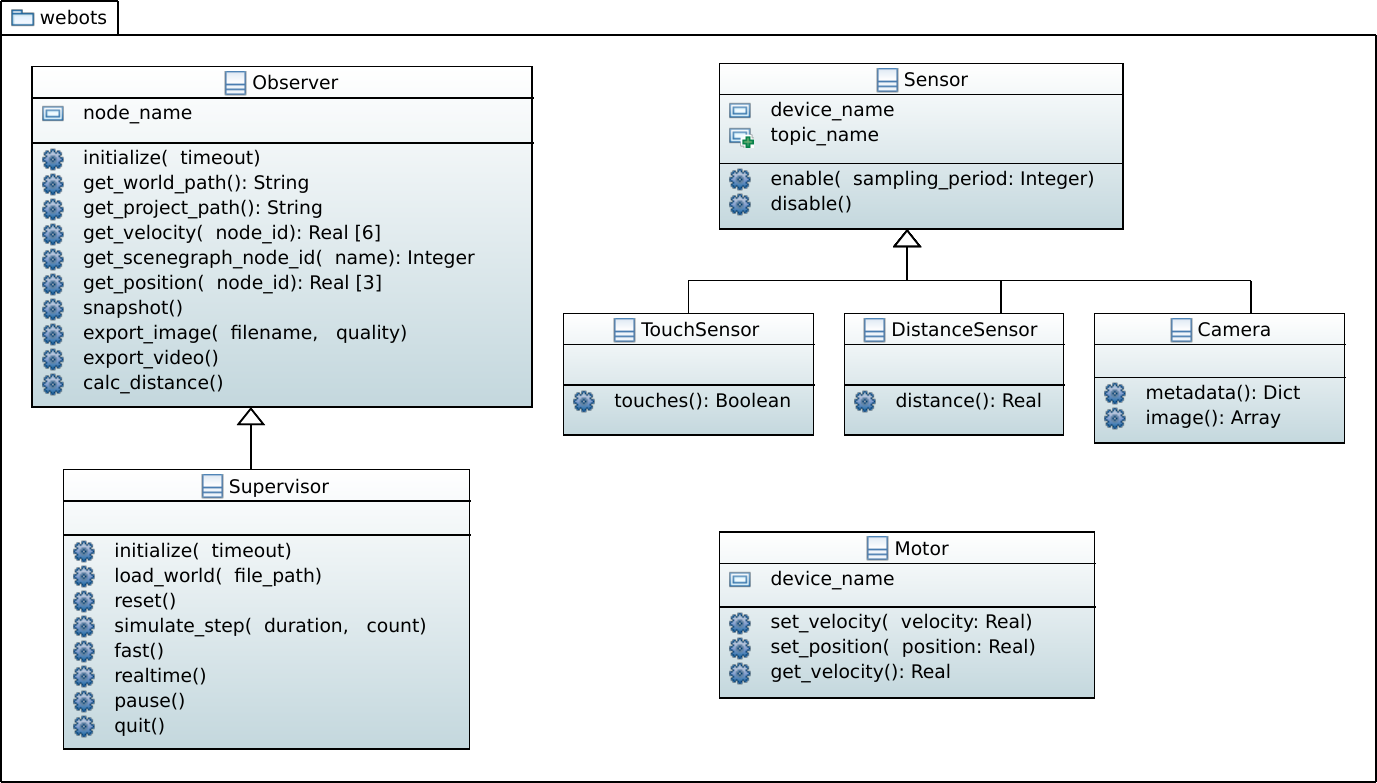}
    \caption{Facade Classes for Observer, Supervisor and Robot Building Blocks} \label{fig:control}
\end{figure}

Most functionality exposed on ros nodes by Webots default ROS controller are services except for sensor values. 
Sensors need to be enabled with a sampling period so that they publish their values on a topic. 
Figure \ref{fig:control} shows the facade classes for touch sensors, distance sensors, cameras and motors. 

The client for the supervisor ROS node is split into two classes: \lstinline{Observer} with probing methods 
that don't affect the simulation and \lstinline{Supervisor} with
altering methods to control the simulation.
Most methods shown for \lstinline{Observer}, \lstinline{Supervisor} and \lstinline{Motor} each use one \lstinline{RosService}. 
The class \lstinline{Sensor} uses a \lstinline{RosService} and a \lstinline{RosSubscriber}.

Note that the classes \lstinline{TouchSensor}, \lstinline{DistanceSensor}, \lstinline{Camera} and \lstinline{Motor} are
not specific to any robot and just represent the client facades for the corresponding Webots concepts\footnote{
    API Functions in Webots Reference Manual: \url{https://cyberbotics.com/doc/reference/nodes-and-api-functions?tab-language=ros}
}. Thus we refer to them as robot building blocks.

\section{Example Application with Robotino}
\label{sec:example}
To apply our approach on an example application we need: a Webots world with the Robotino, a facade to control the Robotino,
a gymnasium environment and a learning algorithm to train an agent.

\subsection{Webots World}
\label{sec:world}
Webots is shipped with several sample worlds and robots\footnote{
    available robot models in Webots: \url{https://webots.cloud/proto?keyword=robot}
} which could serve as starting points for customization. 
A webots world resembles the real world with respect to the task and physical characteristics of the robot. 
The sensor values the robot receives should be the same as in the 
real world. And the actuators should behave the same way.

\begin{figure}[hbt]
    \includegraphics[width=\textwidth]{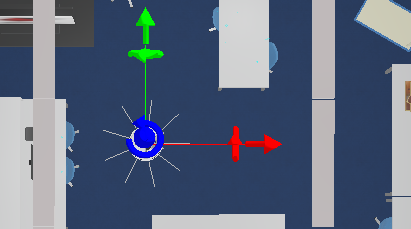}
    \caption{Webots World with Robotino Robot} \label{fig:world}
\end{figure}

We use a simple world with Robotino as robot as shown in figure \ref{fig:world}. 
The task is to move to a given target position and stop there. 
The Robotino can move and collide with objects. The robot has three motors for its omnidrive, nine infrared sensors 
to determine closer distances, a touch sensor to detect collisions, a depth camera and a color camera.

Important settings in the world file are the names for the supervisor and the Robotino that are used as ROS node names as well as 
the timestep that specifies the duration of the physics simulation for one simulation step.

\subsection{Robotino Facade}
\label{sec:robotino}
To interact with the Robotino we create a class according the Facade pattern with the required functionality: obtain sensor 
values and control the motors. We use the robot building blocks for implementation, specifically: 
nine distance sensors, two cameras, a touch sensor and three motors. 
By doing so we have a meaningful and straight 
forward way to interface with the Robotino and the network communication is abstracted away. 

\begin{figure}[htb]
    \includegraphics[width=\textwidth]{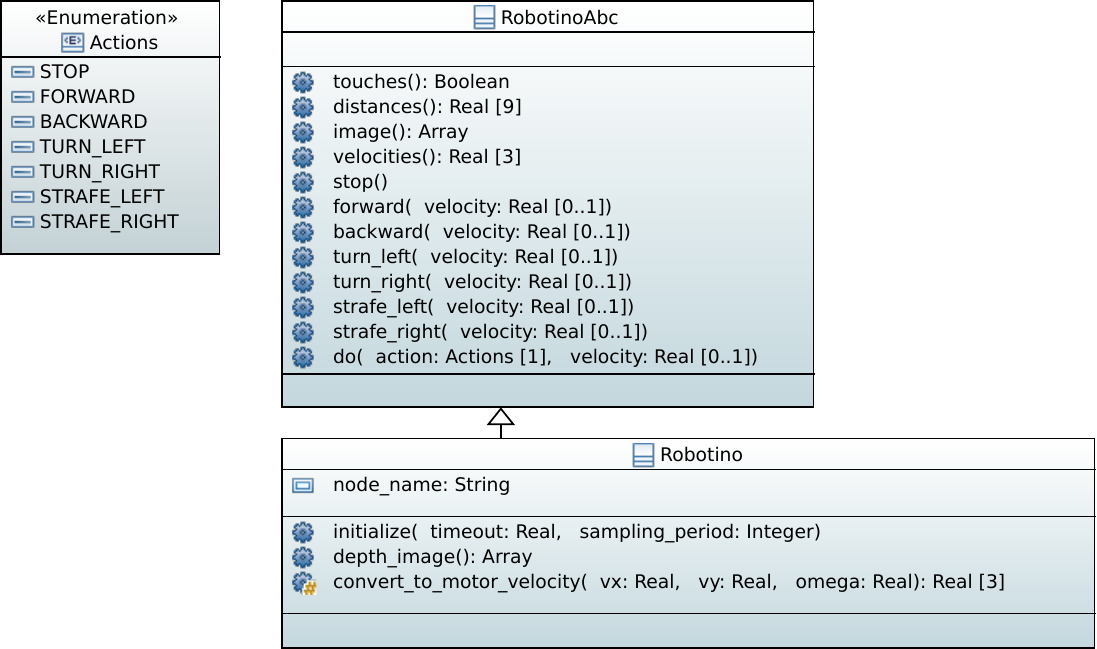}
    \caption{Robotino Interface, Robotino Facade Class} \label{fig:robotino}
\end{figure}

It is important to note that we define an interface \lstinline{RobotinoAbc} first before the realizing facade 
class \lstinline{Robotino} as shown in figure \ref{fig:robotino}. 
There is another realizing class not shown here used for the real Robotino. A common interface for 
both allows us to just switch the implementation to evaluate a trained agent on the real Robotino.

Note that we define the movement actions as enumeration. Some reinforcement learning algorithms require a discrete 
action space. An enumeration is a simple way to map defined literals to integer values.

Compared to approaches without using ROS the class \lstinline{Robotino} is a robot controller usually implemented in C++, Java or 
Python using the Webots API and started by Webots when the user starts the simulation in the graphical user interface.

\subsection{Gymnasium Environment}
\label{sec:env}
Reinforcement learning algorithms rely on an environment which provides the observations and allows agents to perform actions.
The action may change the environment and the agent receives the new observation and a reward. 
A well-established approach for reinforcement learning frameworks to interact with environments is the 
\lstinline{Env} interface of the \lstinline{gymnasium} library, shown in listing \ref{lst:env}.

\begin{lstlisting}[caption={Gymnasium Interface for Environments}, label={lst:env}, captionpos=b, language=Python]
class Env:
    def reset(self): pass
    def step(self, action): pass
    def render(self): pass
    def close(self): pass
\end{lstlisting}

The method \lstinline{reset()} resets the environment to an initial state and returns the initial observation. The method 
\lstinline{step()} is used to perform the given action in the environment. It returns the observation after the action, the 
reward for the action and information about whether and why an episode of steps has ended (e. g. the agent solved the task or 
reached a time limit). Besides the API each environment defines the property \lstinline{action_space} defining all possible 
actions an agent may take, the property \lstinline{observation_space} defining which information the agent receives and 
the reward function.

The observation consists of the sensor values of the Robotino augmented with its position. This information is obtained 
using the classes \lstinline{Supervisor} and \lstinline{Robotino}. During environment reset, the Webots simulation is reset 
and runs until all sensors have valid values.

The environment implements a reward function that assesses the behavior of the agent with respect to its specific task. Based on 
the reward the agent learns to solve the environment. We implemented two reward functions: one that rewards the agent for 
each step taken and another one that rewards the agent only on the last step when the episode ends, i. e. 
the agent solved the environment, had a collision or reached the maximum numbers of steps.

We would like to emphasize that data scientists define the gymnasium environment in their familiar environment for development 
and experimentation. This allows for fast iterations and simple scaling of the agents task to learn. To restrict the degrees of 
freedom for movement of the robot, simply reduce the \lstinline{action_space}. To figure out what difference it makes whether 
the robot has the target coordinates or not, simply adjust the \lstinline{observation_space}. To determine how another 
reward function performs, just change the implementation.

\subsection{Train Agents}
Reinforcement learning algorithms are used to train agents in an environment. 
There are two popular frameworks that provide several agent implementations: \lstinline{tf-agents}\cite{tfagents} based on 
tensorflow and \lstinline{stable baselines3}\cite{stable-baselines3} based on pytorch. 
We evaluate our approach with two agents of the framework \lstinline{tf-agents}:
\begin{description}
    \item[dqn] agent \cite{dqn} as a more recent algorithm that is trained on a fixed number of steps and uses neural networks 
    to predict q-values for actions
    \item[reinforce] agent \cite{reinforce} as a more traditional algorithm that is trained on complete episodes
\end{description}
The algorithms were chosen because of the large time span between their development and their approach: the \lstinline{dqn} agent 
follows a step-centric and the \lstinline{reinforce} agent an episode-centric approach.
Note that the number of steps within episodes may vary.

Listing \ref{lst:setup} outlines how to setup a training session with our approach.
\begin{lstlisting}[caption={Important Steps to Setup a Training Session}, label={lst:setup}, captionpos=b, language=Python]
# create the agent ros node
rospy.init_node('RobotinoAgent', anonymous=False)
# create the webots facade
webots = Webots(supervisor_ros_node_name, world_path)
# create the environment
env_options = {}
tf_env = tf_agents.environments.suite_gym.load(
        'MoveToTargetEnv-v0', gym_kwargs=env_options)
# reset the environment
timestep = webots.run( lambda: tf_env.reset() )
\end{lstlisting}
Agents are typically trained in loops that consist of:
\begin{itemize}
    \item collect experience with the agents policy
    \item train the agent
    \item evaluate the agents policy
\end{itemize}
To prevent Webots from wasting compute resources (see section \ref{sec:webots.facade}) a callback function or 
the \lstinline{with} statement could be used to collect experience or to evaluate a policy. 
Both approaches are illustrated in listing \ref{lst:snippets}.
\begin{lstlisting}[caption={Python Snippets for the Training Loop}, label={lst:snippets}, captionpos=b, language=Python]
# collect experience with callback function
def collect_experience(): pass
webots.run( collect_experience )

# evaluate policy with context manager
with webots:
    # code to evaluate policy
    pass
\end{lstlisting}
During our experiments we discovered a reproducible error in our Webots version: after resetting the simulation for 24 times 
it crashes with a bus error. We addressed that behavior by adjusting our training loop to ensure that Webots is restarted 
after at most 24 simulation resets.

We ran and evaluated a couple of experiments with different hyperparameter settings as well as task difficulties. 
In total more than 100 training sessions with a total duration of more than 200 hours were conducted as shown in 
table \ref{table:summary}. The longest training session lasted over 10 hours.
\vspace{-0.8em}
\begin{table}
\caption{Summary of Experiments}\label{table:summary}
\begin{tabular*}{\textwidth}{@{\extracolsep{\fill} } @{\hspace{3pt}}r@{\hspace{7pt}} | c@{\hspace{7pt}} | r@{\hspace{7pt}} }
Algorithm & Number of Training Sessions & Total Duration  \\
\hline
dqn  &  81  &  132 hours \\
reinforce  &  27  &  83 hours \\
\end{tabular*}
\end{table}
\vspace{-1.5em}

\section{Current Limitation and Workarounds}
\label{sec:limitations}
The current state of implementation has a notable limitation: it is currently not possible to use multiple Webots 
simulation instances per training setup. The hard coded robot names in a Webots world file are used as ROS node names. 
One possibility to address this issue is to use anonymous ROS nodes instead of fixed names and
rely on metadata to identify matching supervisor and robot controllers. 
Another possibility is to preprocess the Webots world file and generate the desired amount of Webots world files for a
particular training session with different fixed names for supervisor and robot ROS nodes. 
The effort seems manageable because the Webots world files are plain text files.
Multiple simulation environments are usually used to accelerate the training process. A common pattern is to use one environment 
to collect experience and another for policy evaluation.

Besides the limitation we already mentioned the two workarounds: first, we switch the Webots simulation modes to prevent 
high cpu usage between simulation steps and second, we design our training loop to restart Webots after at most 24 simulation 
resets to prevent Webots from crashing with a bus error. Both may be related to the version of Webots we used.

Although the current shortcomings might limit the application of the proposed approach, it is important to note that they 
are related to the current state of implementation and not regarding its architecture.

\section{Summary}
\label{sec:summary}
We conducted a couple of experiments with a total training duration of more than 200 hours. 
The longest training sessions took more than 10 hours and involved over 2,500 episodes. 
During those sessions Webots was frequently reset and restarted. 
Despite the implementation workarounds with Webots those training sessions ran robust. 

The presented approach is technologically packed with docker and ROS besides Webots. 
In addition there are a lot of common error sources like spawning and terminating processes and network communication. 
Especially data scientists who focus on algorithm research or task solving don't want to interact with those 
technologies and error sources directly. Instead they demand easy to use APIs which integrate well in their model development 
environment. All settings belonging to their domain should be changeable without switching the working environment to 
ensure a high level of autonomy and productivity. 
In that regard our approach provides a clear separation of responsibilities: 
the creators of virtual worlds together with subject matter experts work in their familiar 
simulation environment on the 3D sceneries with robots. 
Software engineers work with subject matter experts and provide easy to use APIs to control the simulation software and the 
virtual robots and ensure the portability to real robots. 
An infrastructure team bundles the pieces and provides container images. 
Data scientists use them to design gymnasium environments and experiment with different action and 
observation spaces and reward functions to train agents that solve tasks. 
Most notable they don't require knowledge about the simulation software or communication mechanisms used.

The presented results outline that it is practicable to use a standalone simulation environment like Webots in fully 
automated and unattended reinforcement learning setups. Typical use cases include job processing in 
continuous integration systems, batch processing systems or workload managers.
Straight forward Python APIs hide technologies and complexities and allow for robust training sessions. 
Although the Robotino was the main driver for our approach, it is by no means specific to the Robotino. 

Further work addresses the current limitation, more elaborated example applications that require longer training sessions and 
open sourcing the approach.

\subsubsection{Acknowledgements} This work received funding from the Saxony State Ministry for Science, Culture 
and Tourism of the Free State of Saxony, Germany. 

Special thanks go to Tommy Hartmann for his contributions of the Webots world and equipping the virtual Robotino with a 
depth camera.

\bibliographystyle{splncs04}
\bibliography{references.bib}

\end{document}